\documentclass[twoside,twocolumn,9pt]{article}
\usepackage{extsizes}
\usepackage[super,sort&compress,comma]{natbib} 
\usepackage[version=3]{mhchem}
\usepackage[left=1.5cm, right=1.5cm, top=1.785cm, bottom=2.0cm]{geometry}
\usepackage{balance}
\usepackage{mathptmx}
\usepackage{sectsty}
\usepackage{graphicx} 
\usepackage{lastpage}
\usepackage[format=plain,justification=justified,singlelinecheck=false,font={stretch=1.125,small,sf},labelfont=bf,labelsep=space]{caption}
\usepackage{float}
\usepackage{fancyhdr}
\usepackage{fnpos}
\usepackage[english]{babel}
\addto{\captionsenglish}{%
  
}
\usepackage{multirow,booktabs}
\usepackage{array}
\usepackage{droidsans}
\usepackage{charter}
\usepackage[T1]{fontenc}
\usepackage[usenames,dvipsnames]{xcolor}
\usepackage{setspace}
\usepackage[compact]{titlesec}
\usepackage{hyperref}

\usepackage{soul}
\usepackage{xcolor}
\sethlcolor{yellow} 
\usepackage{url}

\newcounter{pat}
\newcounter{pat2}
\usepackage{epstopdf}

\definecolor{cream}{RGB}{222,217,201}

\begin{document}

\pagestyle{fancy}
\thispagestyle{plain}
\fancypagestyle{plain}{
\renewcommand{\headrulewidth}{0pt}
}

\makeFNbottom
\makeatletter
\renewcommand\LARGE{\@setfontsize\LARGE{15pt}{17}}
\renewcommand\Large{\@setfontsize\Large{12pt}{14}}
\renewcommand\large{\@setfontsize\large{10pt}{12}}
\renewcommand\footnotesize{\@setfontsize\footnotesize{7pt}{10}}
\makeatother

\renewcommand{\thefootnote}{\fnsymbol{footnote}}
\renewcommand\footnoterule{\vspace*{1pt}%
\color{cream}\hrule width 3.5in height 0.4pt \color{black}\vspace*{5pt}} 
\setcounter{secnumdepth}{5}

\makeatletter 
\renewcommand\@biblabel[1]{#1}            
\renewcommand\@makefntext[1]%
{\noindent\makebox[0pt][r]{\@thefnmark\,}#1}
\makeatother 
\renewcommand{\figurename}{\small{Fig.}~}
\sectionfont{\sffamily\Large}
\subsectionfont{\normalsize}
\subsubsectionfont{\bf}
\setstretch{1.125} 
\setlength{\skip\footins}{0.8cm}
\setlength{\footnotesep}{0.25cm}
\setlength{\jot}{10pt}
\titlespacing*{\section}{0pt}{4pt}{4pt}
\titlespacing*{\subsection}{0pt}{15pt}{1pt}

\makeatletter 
\newlength{\figrulesep} 
\setlength{\figrulesep}{0.5\textfloatsep} 

\newcommand{\topfigrule}{\vspace*{-1pt}%
\noindent{\color{cream}\rule[-\figrulesep]{\columnwidth}{1.5pt}} }

\newcommand{\botfigrule}{\vspace*{-2pt}%
\noindent{\color{cream}\rule[\figrulesep]{\columnwidth}{1.5pt}} }

\newcommand{\dblfigrule}{\vspace*{-1pt}%
\noindent{\color{cream}\rule[-\figrulesep]{\textwidth}{1.5pt}} }

\makeatother

\twocolumn[
  \begin{@twocolumnfalse}
{\includegraphics[height=30pt]{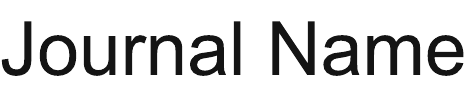}\hfill\raisebox{0pt}[0pt][0pt]{\includegraphics[height=55pt]{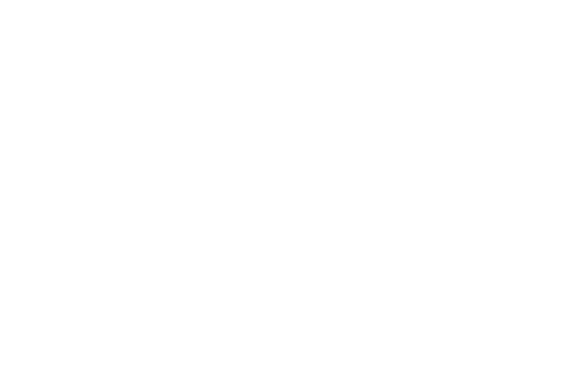}}\\[1ex]
\includegraphics[width=18.5cm]{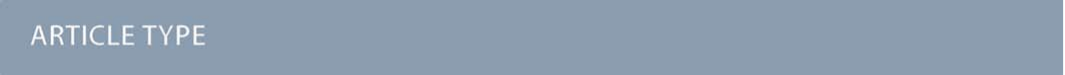}}\par
\vspace{1em}
\sffamily
\begin{tabular}{m{4.5cm} p{13.5cm} }

\includegraphics{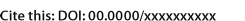} & \noindent\LARGE{\textbf{An AI-enabled tool for quantifying overlapping red blood cell sickling dynamics in microfluidic assays$^\dag$}} \\
\vspace{0.3cm} & \vspace{0.3cm} \\


& \noindent\large{
Nikhil Kadivar,\textit{$^{a\ddag}$}
Guansheng Li,\textit{$^{b\ddag}$}
Jianlu Zheng,\textit{$^{c\ddag}$}
Ming Dao,$^{\ast}$\textit{$^{c}$}
George Em Karniadakis,$^{\ast}$\textit{$^{b}$}
and Mengjia Xu$^{\ast}$\textit{$^{d}$}
} \\

\includegraphics{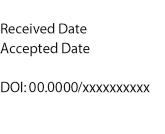} & \noindent\normalsize{Understanding sickle cell dynamics requires accurate identification of morphological transitions under diverse biophysical conditions, particularly in densely packed and overlapping cell populations. Here, we present an automated deep learning framework that integrates AI-assisted annotation, segmentation, classification, and instance counting to quantify red blood cell (RBC) populations across varying density regimes in time-lapse microscopy data. Experimental images were annotated using the Roboflow platform to generate labeled dataset for training an nnU-Net segmentation model. The trained network enables prediction of the temporal evolution of the sickle cell fraction, while a watershed algorithm resolves overlapping cells to enhance quantification accuracy. Despite requiring only a limited amount of labeled data for training, the framework achieves high segmentation performance, effectively addressing challenges associated with scarce manual annotations and cell overlap. By quantitatively tracking dynamic changes in RBC morphology, this approach can more than double the experimental throughput via densely packed cell suspensions, capture drug-dependent sickling behavior, and reveal distinct mechanobiological signatures of cellular morphological evolution. Overall, this AI-driven framework establishes a scalable and reproducible computational platform for investigating cellular biomechanics and assessing therapeutic efficacy in microphysiological systems.} \\

\end{tabular}

 \end{@twocolumnfalse} \vspace{0.6cm}

  ]

\renewcommand*\rmdefault{bch}\normalfont\upshape
\rmfamily
\section*{}
\vspace{-1cm}


\footnotetext{\textit{$^{a}$~School of Engineering, Brown University, Providence, RI, United States}}
\footnotetext{\textit{$^{b}$~Division of Applied Mathematics, Brown University, Providence, RI, United States}}
\footnotetext{\textit{$^{c}$~Department of Materials Science and Engineering, Massachusetts Institute of Technology, Cambridge, MA 02139, United States}}

\footnotetext{\textit{$^{d}$~Department of Data Science, Ying Wu College of Computing, New Jersey Institute of Technology, Newark, NJ, United States}}


\footnotetext[3]{These authors contributed equally to this work.}

\footnotetext[1]{Corresponding authors. E-mail: george\_karniadakis@brown.edu; mingdao@mit.edu; mengjia.xu@njit.edu}

\footnotetext[2]{Electronic Supplementary Information (ESI) available: additional figures and analysis for watershed post-processing.}


\section*{Introduction}

Sickle cell disease (SCD) is a hereditary hemoglobinopathy characterized by the polymerization of hemoglobin S under deoxygenated conditions, leading to abnormal red blood cell (RBC) shape changes and altered microcirculatory flow~\cite{munoz1997folding,eaton1990sickle,ye2019red,li2023silico,li2023combined,dorken2023circulating}. The morphological transformation of RBCs—from biconcave discocytes to sickled shapes—plays a critical role in the pathophysiology of vaso-occlusion, hemolysis, and impaired oxygen transport~\cite{li2020numerical,sundd2019pathophysiology,li2024red}. Quantitative analysis of these morphological transitions is therefore central to understanding the biophysical mechanisms driving disease progression and evaluating therapeutic interventions.

Although experimental imaging and microfluidic assays have advanced considerably, achieving automated and quantitative classification of sickle cell morphologies continues to pose major analytical challenges~\cite{xu2017deep,qiang2024framework,alzubaidi2020deep,darrin2023classification,wang2021margination}. Conventional image analysis methods often rely on manual feature extraction, thresholding, or heuristic shape metrics that lack robustness across heterogeneous patient samples and varying imaging conditions~\cite{carvalho2021manual,dorken2023circulating,tavakoli2021new,kiu2022geometric}. Moreover, such approaches rarely capture the temporal evolution of cell morphology under dynamically changing biophysical environments, such as drug, oxygen level, and cell overlap~\cite{recktenwald2022red,alapan2016dynamic,li2023analysis}. Recent advances in deep learning and image segmentation, particularly convolutional neural network (CNN)–based architectures, have revolutionized biomedical image analysis by enabling data-driven extraction of morphological and contextual features beyond traditional handcrafted descriptors~\cite{litjens2017survey,minaee2021image}. CNNs have emerged as powerful tools for extracting biophysical and mechanistic information directly from biomedical imaging data, bridging the gap between visual pattern recognition and physiological interpretation~\cite{qiang2024framework,liu2021review,rayed2024deep,mehdi2025towards, neelakantan2025artificial}. Among these, the U-Net architecture and its derivatives have become foundational in biomedical imaging owing to their encoder–decoder symmetry and skip connections, which facilitate precise localization of fine cellular boundaries while preserving global contextual information~\cite{ronneberger2015u,zhou2018unet++,zhang2022aoslo,mehdi2025non}. The nnU-Net framework, in particular, has demonstrated exceptional adaptability and generalization across diverse biomedical datasets by automatically optimizing network configurations and preprocessing pipelines~\cite{isensee2021nnu}. 

\begin{figure*}[!htbp]
    \centering
    \includegraphics[width=1.0\linewidth]{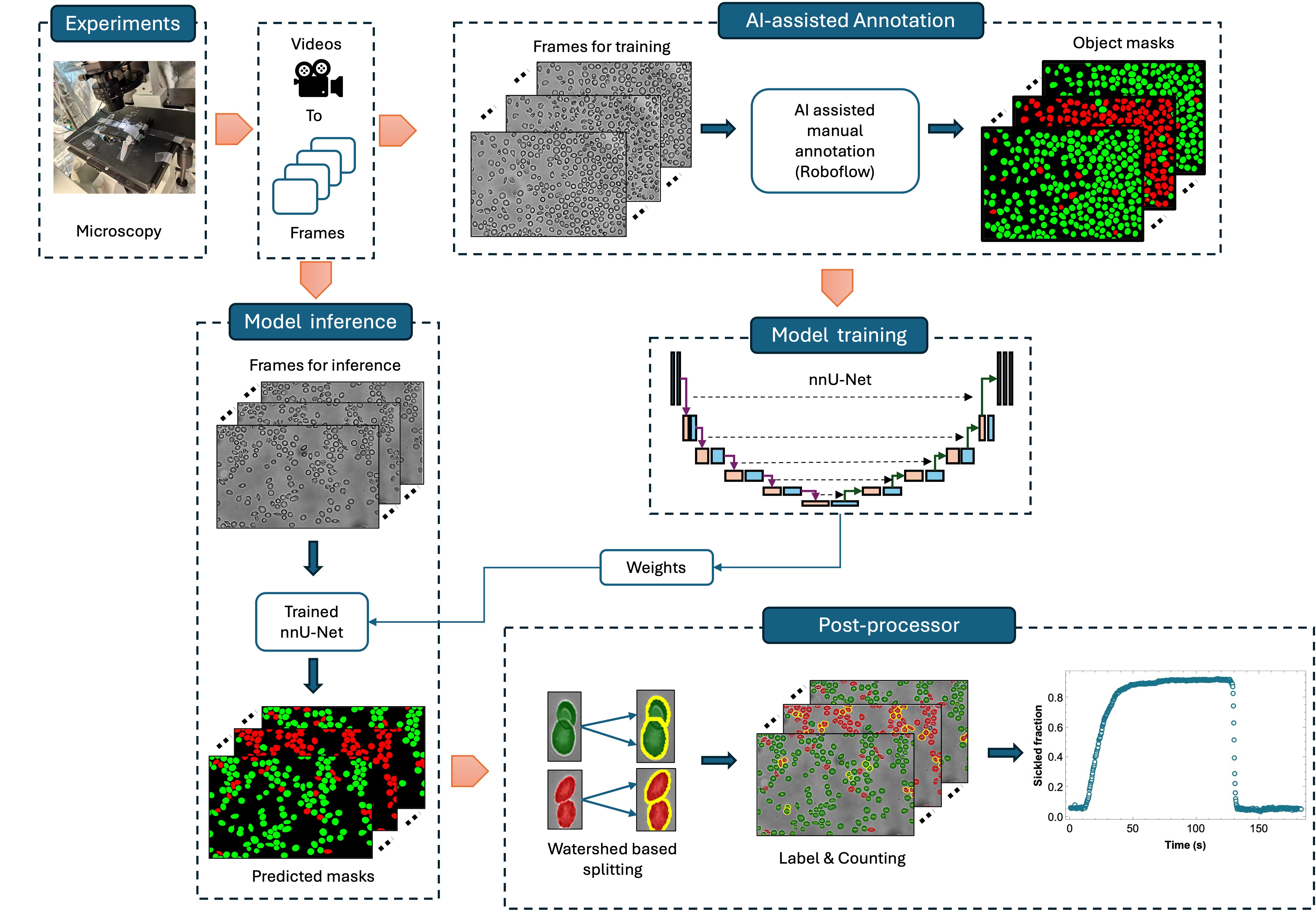}
    \caption{Schematic representation of the AI-enhanced segmentation framework for quantifying RBC sickling dynamics in dense and overlapping fields. A subset of images from microfluidic experiments is sampled to obtain representative frames at user-defined intervals. These frames are subsequently annotated in Roboflow using AI-assisted manual labeling to generate instance masks of red blood cells (RBCs) labeled  as healthy or sickled. The annotated images are used to train an enhanced nnU-Net segmentation model that automatically optimizes preprocessing and network configurations. The optimized model weights are then applied during inference to produce predicted segmentation masks for unseen experimental datasets. A marker-controlled watershed post-processing step refines and separates overlapping cells, enabling accurate instance counting and classification. Finally, the temporal evolution of the sickled fraction is quantified from videos of sickling dynamics.}
    \label{fig:Figure1}
\end{figure*}

Despite these advances, the application of deep learning frameworks to dynamic biophysical processes—such as red blood cell (RBC) sickling dynamics and population-level morphological evolution—remains limited, particularly in addressing challenges like overlapping cells and accurate quantification within dense suspensions. The development of robust computational frameworks capable of resolving such overlaps is therefore essential, both for reliably quantifying sickling and unsickling dynamics and for enabling higher-throughput experiments via densely packed cell suspensions. In this study, we present an automated deep learning framework for quantifying red blood cell (RBC) sickling dynamics from experimental data~(Figure~\ref{fig:Figure1}). The workflow integrates AI-assisted annotation in Roboflow with an enhanced nnU-Net architecture, followed by a watershed algorithm to achieve robust segmentation and classification of RBC morphologies. The model was trained on heterogeneous datasets derived from sickle cell disease (SCD) patient samples, facilitating accurate identification and temporal characterization of red blood cell morphological dynamics. By integrating image-based segmentation with quantitative analysis of temporal morphological evolution, the framework effectively captures the progression of sickling driven by oxygen-dependent biophysical alterations, offering new insights into the mechanobiological mechanisms underlying RBC shape evolution. This AI-driven platform provides a scalable and reproducible computational tool for objective evaluation of cellular biomechanics and therapeutic responses within microphysiological systems.

\section*{Materials and methods}

\textbf{Preparation of RBC suspensions.} Silicone elastomer base and curing agent (Sylgard 184) were obtained from Dow Chemical Company. Whole blood samples were collected from homozygous SCD patients at Massachusetts General Hospital under an Excess Human Material Protocol approved by the Partners HealthCare Institutional Review Board (IRB), with a waiver of informed consent. Following the pretreatment procedure previously described by our group~\cite{du2015kinetics}, packed red blood cells (RBCs) were gently washed three times with phosphate-buffered saline (1× PBS; Sigma-Aldrich, St. Louis, MO, USA) by centrifugation at 1,500 rpm for 3~min at room temperature. The washed RBCs were resuspended in PBS containing 1\% (w/v) bovine serum albumin (BSA; EMD Millipore, Billerica, MA, USA) to achieve a hematocrit of 2\%.  

\textbf{Double-Layer Microfluidic Device.} Microfluidic devices were fabricated following previously reported methods~\cite{qiang2024framework}. Briefly, polydimethylsiloxane (PDMS) was prepared by mixing the elastomer base and curing agent in a 10:1 (w/w) ratio and curing the mixture overnight at 80~\textdegree C. Two PDMS layers—one forming the gas channel and the other the cell channel—were cast from silicon wafer molds and subsequently bonded to form a double-layer configuration.

\textbf{Sickling Kinetics Assay.} Centrifuged RBCs were treated with osivelotor (formerly known as GBT021601; Pfizer, New York, NY, USA) at 0\% (no treatment) and 100\% modification levels, based on a 1:1 molar ratio of osivelotor to total hemoglobin. The treated suspensions were incubated at room temperature for 1~h and stored at 4~\textdegree C until imaging. Brightfield videos were recorded using a high-resolution CMOS camera (The Imaging Source, Charlotte, NC, USA) mounted on an Olympus X71 inverted microscope (Olympus America, Breinigsville, PA, USA) equipped with a 60× oil-immersion objective lens (NA = 1.25). Recordings were acquired under ambient conditions at 4~frames/s with a resolution of 5472~×~3648~(RGB64). To induce hypoxia, a gas mixture of 2\% O\textsubscript{2} and 5\% CO\textsubscript{2} balanced with N\textsubscript{2} was introduced into the upper gas channel intersecting the lower cell channel. The RBC suspension (2\% hematocrit) was prepared after centrifugation at 1,500~rpm and triple washing with PBS.  

\section*{Data annotation and labeling}

\begin{figure*}[!htbp]
    \centering
    \includegraphics[width=1.0\linewidth]{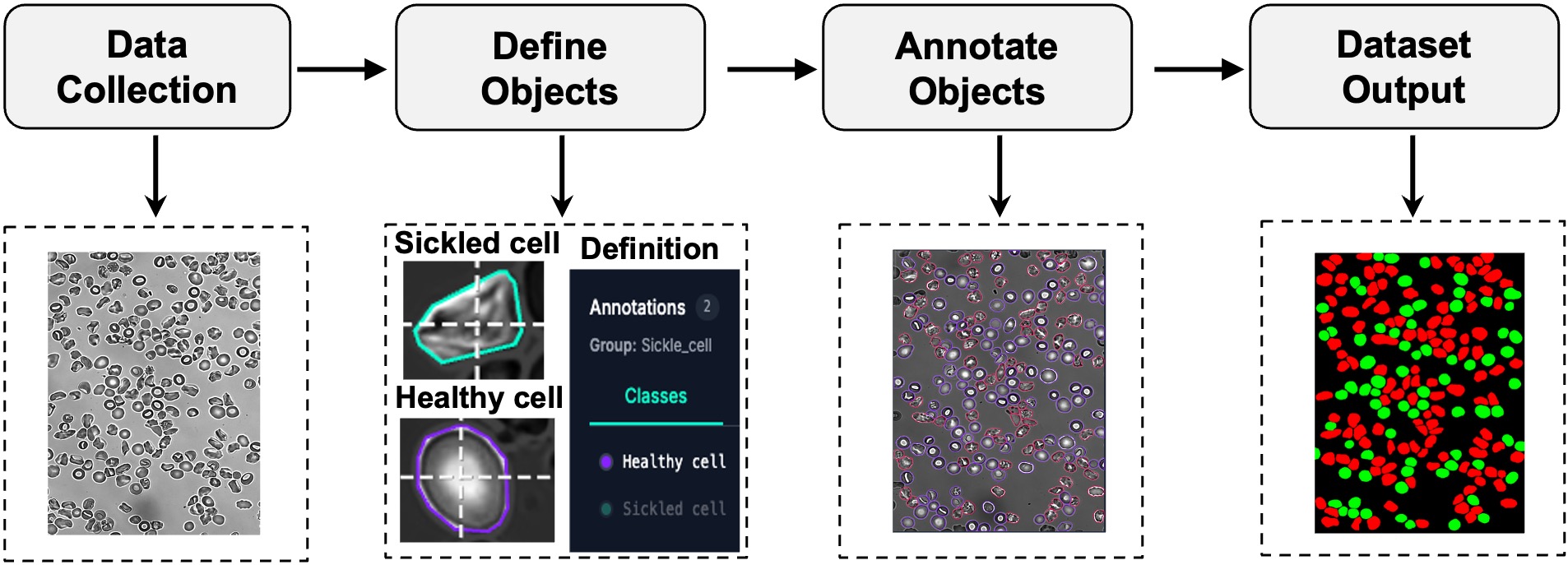}
    \caption{Workflow for dataset acquisition and annotation using the Roboflow platform~\cite{dwyer2024roboflow}. Starting from microscopy frames (Data Collection), cell classes are defined (Healthy and Sickled) and used to guide annotation in Roboflow via AI-assisted manual labeling. The resulting labeled dataset (Dataset Output) provides class-specific masks for downstream model training and evaluation, with background in black, healthy cells in green, and sickled cells in red.}
    \label{fig:Figure2}
\end{figure*}

Before performing sickle cell classification on the experimental videos, a representative subset of 12 image frames was annotated to construct the training dataset for the nnU-Net segmentation model. The Roboflow platform~\cite{dwyer2024roboflow} was employed for data annotation and labeling, providing an efficient interface for segmentation mask generation and dataset management. Figure~\ref{fig:Figure2} illustrates the annotation workflow implemented in Roboflow.

\textbf{Data collection.} First, 12 representative image frames were extracted from the experimental videos and uploaded to the annotation platform. Owing to variability in experimental conditions—including cell population density, cellular morphology, sickling duration, and imaging quality—the selected frames exhibited substantial heterogeneity. 

\textbf{Object definition and annotation.} Annotation categories were defined for three classes: healthy cells, sickle cells, and background. Creating labeled datasets for cellular segmentation is typically labor-intensive and time-consuming. To streamline this process, AI-assisted annotation tools, including the Segment Anything Model (SAM 2)~\cite{ravi2024sam} integrated within Roboflow, were utilized to facilitate rapid and consistent mask generation across diverse imaging conditions. Using a semi-automated workflow, the annotator selected individual target cells, after which the SAM-based segmentation assistant automatically generated precise masks delineating cell boundaries. Each red blood cell was subsequently reviewed and assigned to the appropriate class to ensure comprehensive and accurate classification. The resulting masks captured detailed RBC morphologies across representative frames spanning multiple time points across a few experiments.

\textbf{Dataset output.} The finalized annotations were exported as per-pixel label maps and converted into the nnU-Net data format, with label indices defined as background = 0, healthy = 1, and sickled = 2.
Although the present work focuses on two morphological classes of RBCs, the same annotation and training framework can be readily extended to additional morphological or cellular classes for broader biological imaging applications.

\section*{Training and inference}

To enable robust and reproducible segmentation of the RBCs, the training dataset was augmented using random flips and rotations, preserving cellular morphology while improving robustness to orientation variability. As a preprocessing step, all frames were converted to grayscale and contrast-normalized using contrast-limited adaptive histogram equalization (CLAHE), implemented via OpenCV~\cite{bradski2000opencv}, to mitigate illumination non-uniformity in microfluidic recordings. Building on this standardized training set, we employed the nnU-Net framework, a self-configuring deep learning system that automatically adapts preprocessing, architecture, and training hyperparameters to a given biomedical dataset~\cite{isensee2021nnu}. nnU-Net builds upon the well-established U-Net architecture~\cite{ronneberger2015u} but extends it with dynamic configuration of parameters such as input normalization, patch size, and network depth without requiring manual tuning. This adaptability makes it particularly suited for biomedical imaging tasks where dataset size, contrast, and scale can vary substantially.

\textbf{Training.} We trained an nnU-Net segmentation model in a 2D configuration using 5-fold cross-validation to segment experimental microscopy frames into healthy and sickle RBC classes. In each fold, four folds were used for training and one for validation, producing fold-specific model checkpoints that can be combined as an ensemble at inference. Despite the small training set (12 annotated frames), the model achieved strong segmentation performance, underscoring the efficiency of the AI-assisted labeling workflow and the generalization capability of nnU-Net for our biophysical imaging task. Unless otherwise specified, we used the default nnU-Net v2 training configuration~\cite{isensee2021nnu}, which automatically infers key preprocessing and training parameters (e.g., network depth, patch size, batch size, and sampling strategy) from the dataset properties and available GPU/CPU-memory. Training followed the standard nnU-Net training recipe, using stochastic gradient descent with Nesterov momentum and a polynomial learning-rate schedule, together with a composite Dice + cross-entropy loss. Inputs were Z-score normalized on a per-image basis, and patches were sampled with foreground oversampling to ensure adequate representation of RBC pixels. The best model checkpoint for each fold was selected automatically based on validation score.

\textbf{Inference.} After training, we developed a standardized nnU-Net inference pipeline to segment healthy and sickle RBCs in microscopy frames extracted from experimental videos. These frames may correspond to time points not included in training or may originate from new microfluidic experiments not represented in the annotated dataset. The inference workflow converts experiment videos into nnU-Net--compatible inputs and produces per-frame label maps efficiently and reproducibly. The procedure is summarized below.

\begin{enumerate}
    \item Frame extraction and input standardization: Raw experimental videos were decoded using OpenCV (cv2) and converted into individual image frames. During this step, frames were preprocessed (grayscale conversion and CLAHE-based contrast normalization) and resized to match the training resolution (\(1000 \times 1000\) pixels for the trained model). Frames were then saved in the nnU-Net format using the required naming convention, ensuring direct compatibility with nnU-Net inference.

    \item Model inference: The trained nnU-Net model was subsequently applied to the extracted frames to generate segmentation masks identifying healthy and sickled RBCs.

    \item Output generation: Predicted segmentations were exported as PNG label maps with integer encodings corresponding to background (0), healthy (1), and sickle (2) regions, providing standardized outputs for downstream quantitative and morphological analysis.

\end{enumerate}

Inference was executed using GPU hardware, enabling fast batch processing of several video datasets with reproducible outputs. Before inference, we resized every frame to the same spatial resolution used during training (\(1000 \times 1000\) pixels);empirically, this improved generalization accuracy and was therefore adopted as the default preprocessing setting. The predicted masks were subsequently post-processed for cell counting and sickled fraction estimation as described in the next section.

\section*{Watershed-based separation of overlapping cells}

 In dense suspensions, touching or overlapping RBCs are frequently merged into a single connected region within a class mask, which biases downstream quantification. To improve instance-level quantification, we implemented a watershed method to separate overlapping cells within each class mask~\cite{beucher2018morphological, van2014scikit}. Importantly, nnU-Net provides reliable segmentation of overlapping cells, including heterogeneous assemblies of healthy and sickle cells. Homogeneous overlapping cells are subsequently separated using a watershed technique.

\begin{figure*}[!htbp]
    \centering
    \includegraphics[width=1.0\linewidth]{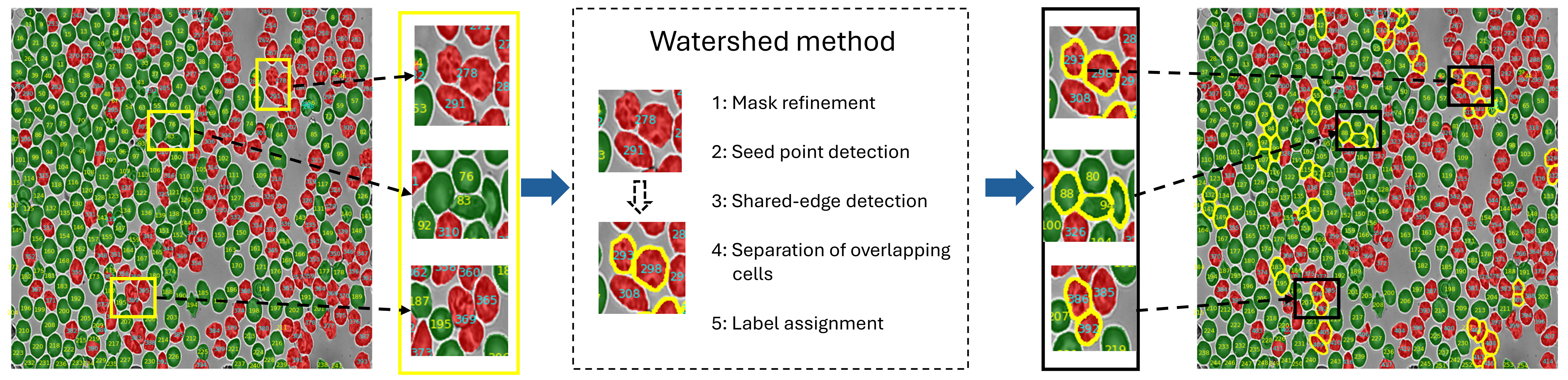}
    \caption{Overview of the watershed pipeline for separating overlapping RBCs. The left panel shows the nnU-Net segmentation result, where overlapping same-class cells may appear as merged regions. The middle panel summarizes the marker-controlled watershed steps, and the right panel shows the instance-separated output after watershed. Green indicates healthy cells, red indicates sickle cells, and yellow outlines highlight regions that were successfully split.}
    \label{fig:Figure3}
\end{figure*}

 Figure~\ref{fig:Figure3} illustrates the complete workflow of the watershed method to separate overlapping cells. The left panel shows the segmentation results obtained from the nnU-Net prediction, in which healthy cells (green) and sickle cells (red) are masked. The yellow bounding box highlights an example of overlapping cells identified in the prediction, together with a zoomed-in view for clarity. The middle panel presents the key steps of the watershed-based separation procedure. \textbf{Mask refinement.} For each frame, the nnU-Net label map was first converted into class-specific binary masks corresponding to healthy and sickle red blood cells (RBCs). Small spurious regions were removed using an area-based connected-component filter with a minimum object-size threshold, thereby suppressing segmentation noise while preserving valid cell regions. \textbf{Seed point detection.} The overlapping-cell separation procedure began by computing a distance transform within each class-specific mask to identify approximate cell centers, which served as seed points (markers) for boundary propagation. 
A smoothed distance map was generated from each binary mask, where each individual cell center appears as a local maximum. \textbf{Shared-edge detection.} These local maxima act as markers from which virtual boundaries expand outward until neighboring regions meet, corresponding to the shared edges of overlapping cells and thereby delineating their common boundaries. To control marker generation, these local maxima were retained only if their distance-transform values exceeded a relative peak-height threshold (set as a fraction of the maximum distance value within the merged region). Additionally, a minimum inter-peak distance constraint was enforced to suppress closely spaced maxima and prevent over-segmentation within a single cell. For completeness, the ESI$^{\dag}$ provides a sensitivity analysis of these marker-generation hyperparameters and identifies a stable operating range that balances under- and over-segmentation in dense suspensions. \textbf{Separation of overlapping cells.} The overlapping cells were subsequently split along the detected boundaries, dividing the merged region into distinct cell instances.  \textbf{Label assignment.} Finally, each detected cell was assigned a unique label, enabling robust cell-by-cell quantification across densely packed experimental frames. 
The right panel of Figure~\ref{fig:Figure3} shows the final segmentation results, demonstrating successful separation of overlapping cells.









\textbf{Quantitative outputs and sickled fraction.} For each frame, the pipeline reports the number of healthy and sickled cells. The resulting sickled fraction for each frame at each time point is given by,
\begin{equation}
\label{eq:sickling-fraction}
\mathrm{Sickled\ Fraction} \;=\; \frac{n_{\mathrm{sickled}}}{n_{\mathrm{healthy}} + n_{\mathrm{sickled}}}\,,
\end{equation}
where $n_{\mathrm{healthy}}$ and $n_{\mathrm{sickled}}$ denote the per-frame counts of healthy and sickle RBC instances, respectively.
In addition to tabular outputs, per-frame overlays are generated and saved for quality control, enabling visual verification of segmentation results and watershed boundaries in crowded regions.

\section*{Limitations of existing microfluidic sickling quantification approaches}

Accurate quantification of sickling dynamics is critical for drug screening, mechanobiology studies, and high-throughput assay development~\cite{costa2016sickle, metaferia2022phenotypic}. However, microfluidic experiments are typically performed under dense and overlapping cell conditions, in which several commonly used analysis approaches become unreliable or impractical~\cite{alapan2016dynamic, guo2014microfluidic}. Manual counting can be feasible for sparse fields of view, but it quickly becomes time-consuming, subjective, and increasingly error-prone as cell density increases, severely limiting assay throughput and hindering systematic studies such as multi-condition drug screening or repeated trials~\cite{caicedo2019evaluation}. Shape-descriptor methods, which summarize connected regions using global geometric features and implicitly assume a one-object–one-shape relationship, break down under overlap: multiple cells merge into a single connected component with no visible internal boundaries, leading to loss of object identity and unreliable instance separation~\cite{meijering2012cell, ulman2017objective}. Threshold- or intensity-based methods rely on intensity contrast to define object boundaries, but overlapping cells often exhibit continuous or additive intensity profiles that eliminate contrast at contact regions, causing multiple cells to merge into a single foreground region~\cite{wan2019accurate}. Collectively, these limitations motivate the development of segmentation and quantification methods that (i) remain robust under dense and overlapping conditions, (ii) provide instance-level identification and counting rather than region-level segmentation alone, and (iii) enable reproducible, time-resolved analysis suitable for high-throughput microfluidic assays~\cite{stringer2021cellpose, moen2019deep}.

\section*{Temporal evolution of the sickle cell fraction across different cell suspension densities}

\begin{figure*}[!htbp]
    \centering
    \includegraphics[width=0.9\linewidth]{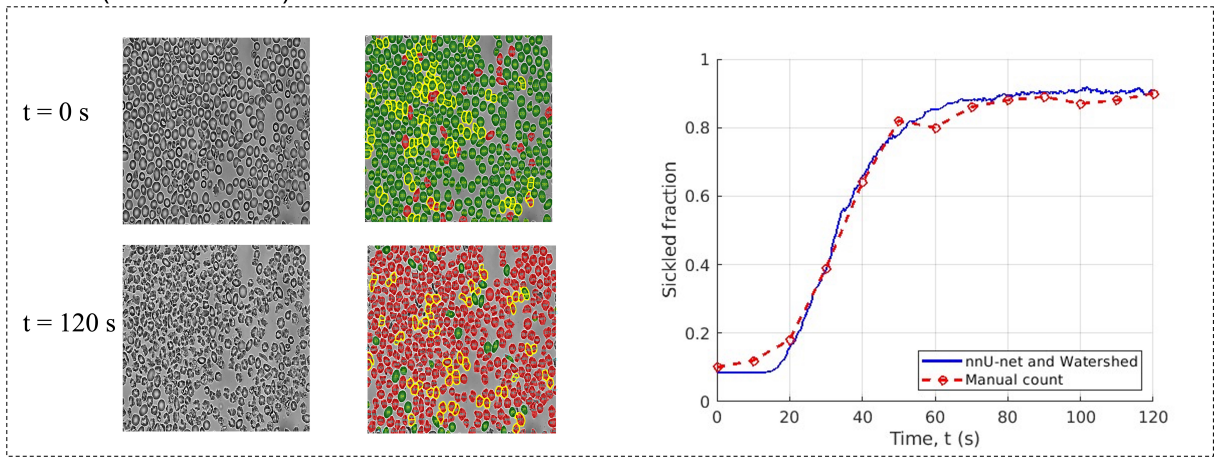}
    \caption{(A--C) Representative sickling dynamics at different cell suspension densities from the same patient, shown at $t = 0$~s (top row) and $t = 120$~s (bottom row). 
For each condition, the left column shows the original micrograph, the middle column displays the nnU-Net prediction with watershed-based instance separation (green: healthy; red: sickle; yellow: overlapping regions separated by watershed), and the right column compares the predicted sickle-cell fraction (solid blue) with manual counts (red dashed).
}
    \label{fig:Figure4}
\end{figure*}

To evaluate the predictive accuracy of the proposed framework, we performed controlled experiments under varying cell suspension densities. The corresponding patient sample parameters are summarized in Table~\ref{tab:SickledExp} Sample~\Roman{pat}. Using the developed approach, we quantified the time-dependent evolution of the sickle-cell fraction across different suspension densities, as shown in Figure~\ref{fig:Figure4}. The cell suspension density was defined by the total number of cells detected in each video frame (frame resolution: $5472 \times 3648$~pixels). Representative cases with 62 (Video 1), 170 (Video 2), and 417 (Video 3) cells are illustrated in Figure~\ref{fig:Figure4}(A-C), respectively. For each density level, the left column shows the original image and the middle column shows the segmentation overlay after nnU-Net and watershed-based instance separation; the top and bottom rows correspond to $t=0$~s and $t=120$~s, respectively. In the overlays, green denotes healthy cells, red denotes sickle cells, and yellow highlights overlapping regions that are separated by the watershed post-processing. The rightmost panel illustrates the temporal evolution of sickling dynamics under different suspension densities. The red dashed curves represent manually quantified results and the solid curves denote predictions generated by the combined nnU-Net and watershed framework. The mean absolute error (MAE) in sickle-cell fraction was 0.032, 0.040, and 0.023 for cases 1–3, respectively. While Figure~\ref{fig:Figure4} confirms accurate recovery of the sickle-cell fraction over time, Figure~S1 (ESI$^{\dag}$) further demonstrates that watershed-based instance separation improves cell counting accuracy compared with nnU-Net segmentation alone. Overall, the sickling dynamics predicted by the nnU-Net augmented with watershed post-processing exhibit strong agreement with manual quantification, demonstrating the accuracy and robustness of the proposed method across a wide range of cell suspension densities. To avoid overlapping RBCs that can reduce counting accuracy, experiments typically use less densely packed suspensions (e.g., Figure~\ref{fig:Figure4}B). By allowing the use of  denser, overlapping suspensions (e.g., Figure~\ref{fig:Figure4}C), our method increases the experimental throughput 2.5-fold. 

\section*{Temporal evolution of the sickle cell fraction under different drug treatments }

\begin{figure*}[!htbp]
    \centering
    \includegraphics[width=0.7\linewidth]{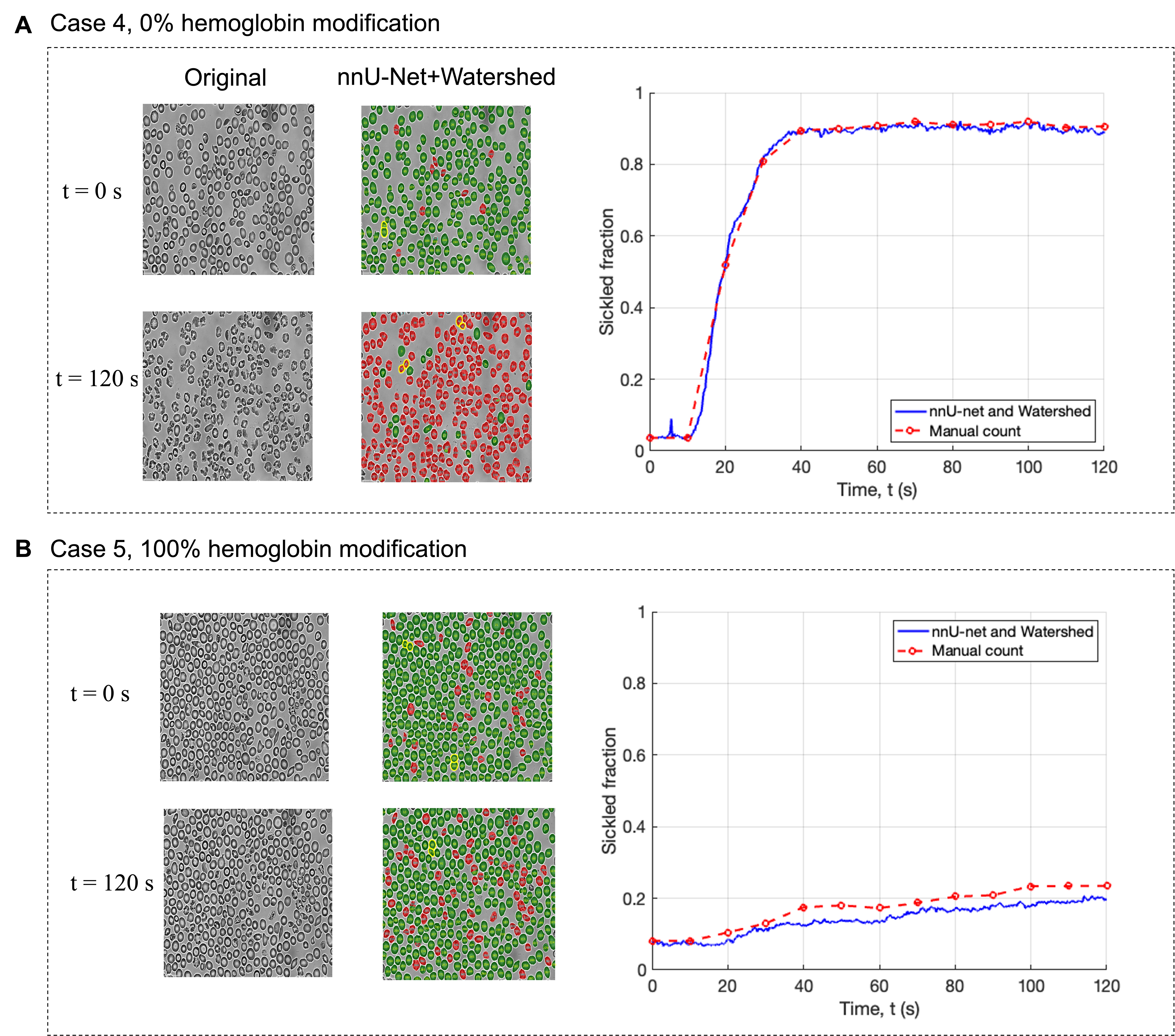}
    \caption{Effect of hemoglobin modification on RBC sickling dynamics. (A,B) Time-resolved micrographs of RBCs under 2\% O$_2$ with (A) 0\% and (B) 100\% hemoglobin modification by osivelotor at $t = 0$~s and $t = 120$~s (left panel). Comparison between predicted and manually counted sickling dynamics under different hemoglobin modification levels (right panel).}

\label{fig:Figure5}
\end{figure*}

We also tested the proposed framework under different hemoglobin modification levels, as shown in Figure~\ref{fig:Figure5}, where the time-resolved snapshots of sickling dynamics observed \textit{in~vitro} under varying degrees of hemoglobin modification by osivelotor (0\% (Video 4) and 100\% (Video 5)). The corresponding patient sample parameters are summarized in Table~\ref{tab:SickledExp} Sample~\Roman{pat2}. At the onset of the experiment ($t=0$~s), the majority of red blood cells (RBCs) exhibited the characteristic biconcave morphology associated with normoxic conditions. As hypoxia (2\% oxygen) progressed over time to $t=120$~s, the RBCs underwent morphological transformations into elongated and crescent-shaped forms, representing various sickling states~\cite{li2023silico}. In contrast, under 100\% hemoglobin modification (Figure~\ref{fig:Figure5}B), the RBC morphology at $t=120$~s remained largely similar to that at $t=0$~s, indicating a substantial inhibition of sickling. Figure~\ref{fig:Figure5}C compares the predicted sickled fraction from our computational method with that obtained by manual counting. Under 0\% hemoglobin modification, the final sickled fractions were approximately 91.0\% (prediction) and 91.8\% (manual), whereas under 100\% modification, the fractions declined markedly to 19.2\% (prediction) and 23.5\% (manual). The close agreement between the two measurements demonstrates that the proposed framework accurately reproduces experimental observations, confirming its robustness and reliability in quantifying sickling dynamics.

{\renewcommand{\arraystretch}{1.5}
\begin{table*}[h]
\caption{
Clinical information about samples in the experiment.
}
\begin{center}
{\scriptsize
\begin{tabular*}{\textwidth}{@{\extracolsep{\fill}}llllll}
\hline

Parameter & $T$ ($^{\circ}\mathrm{C}$) &  $P_{O_2}$ (mmHg)  & MCV ($\mu m^3$) & MCHC (g/dL) & HbS\% \\

\hline
\multicolumn{1}{c}{Sample~\Roman{pat}} &
\multicolumn{1}{c}{25} & 
\multicolumn{1}{c}{15.2} & 
\multicolumn{1}{c}{88.5} & 
\multicolumn{1}{c}{36.8} & 
\multicolumn{1}{c}{86.3\%} \\
\hline

\multicolumn{1}{c}{Sample~\Roman{pat2}} &
\multicolumn{1}{c}{25} & 
\multicolumn{1}{c}{15.2} & 
\multicolumn{1}{c}{84.6} & 
\multicolumn{1}{c}{35.4} & 
\multicolumn{1}{c}{85.4\%} \\
\hline

\end{tabular*}
\label{tab:SickledExp}
}
\end{center}
\end{table*}
}

\section*{Sensitivity analysis of watershed hyperparameters}

\begin{figure*}[!h]
    \centering
    \includegraphics[width=0.9\linewidth]{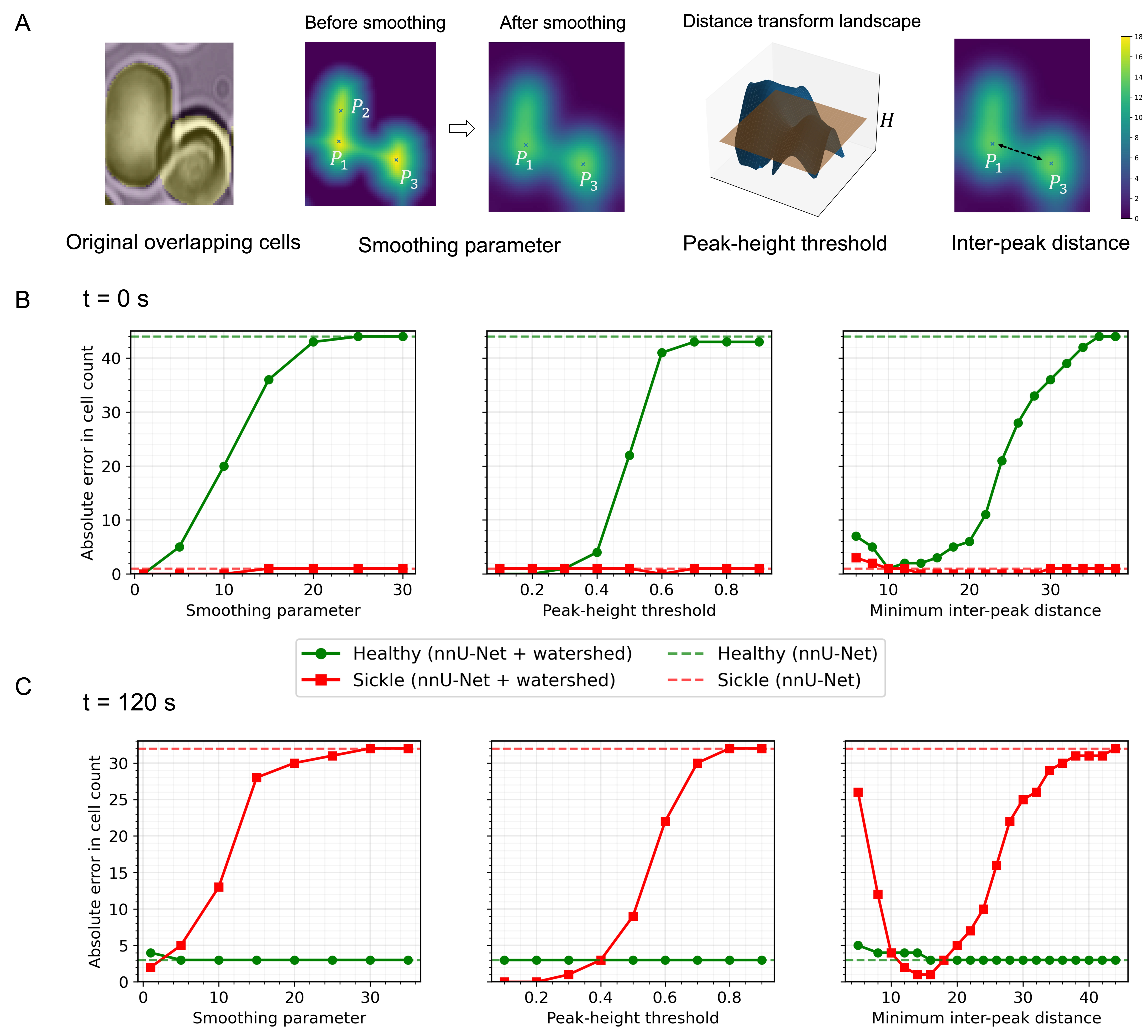}
    \caption{\textbf{Sensitivity analysis of watershed hyperparameters for marker-based instance separation (Case~3, $N = 417$ RBCs)}.
\textbf{(A)} Schematic illustration of the watershed marker-generation process for overlapping cells, including peak detection before and after Gaussian smoothing of the distance transform, application of a peak-height threshold $H$ for marker generation, and enforcement of a minimum inter-peak distance.
\textbf{(B--C)} Absolute cell-counting error relative to manual counting for healthy (green) and sickled (red) cells as key watershed parameters are varied: distance-transform smoothing parameter (left), peak-height threshold for marker generation (middle), and minimum inter-peak distance (right). Results at $t = 0$~s (\textbf{B}) and $t = 120$~s (\textbf{C}) highlight distinct sensitivity regimes for healthy versus sickled RBC populations. Solid curves correspond to the full nnU-Net + watershed pipeline, while dashed horizontal lines indicate the baseline counting error obtained using nnU-Net segmentation without watershed post-processing.}
    \label{fig:S2}
\end{figure*}

To assess the sensitivity of watershed-based instance separation to its key hyperparameters, we performed a systematic analysis of the marker-generation stage using a dense cell suspension with $N = 417$ RBCs (Case~3; Figure~\ref{fig:S2}). The analysis was conducted at two representative time points: $t = 0$~s, when the population is dominated by healthy cells, and $t = 120$~s, when sickled cells are predominant. Figure~\ref{fig:S2}A schematically illustrates the marker-generation process underlying watershed post-processing. First, a distance transform is computed, yielding a scalar field whose local maxima correspond to candidate cell centers. Prior to smoothing, multiple nearby local maxima (e.g., $P_1$, $P_2$, and $P_3$) may arise within partially overlapping or elongated regions due to boundary irregularities and segmentation noise. In this example, $P_1$ and $P_2$ correspond to closely spaced peaks within a single cell, whereas $P_3$ represents a distinct neighboring cell. Gaussian smoothing of the distance transform suppresses minor local extrema and consolidates redundant peaks, preserving dominant maxima (e.g., $P_1$ and $P_3$) while eliminating spurious peaks such as $P_2$. A relative peak-height threshold $H$ is subsequently applied to reject shallow maxima, followed by enforcing a minimum inter-peak distance to suppress closely spaced markers and prevent over-segmentation. Together, these steps transform a noisy distance landscape into a sparse, well-separated set of markers that reliably represent individual cell instances prior to watershed partitioning.

For each hyperparameter sweep (Figure~\ref{fig:S2}B–C), we quantified the absolute cell-counting error relative to manual annotation for healthy and sickled cells separately and compared performance against the nnU-Net–only baseline. Consistent with class prevalence, counting error at $t = 0$~s is dominated by healthy cells, whereas at $t = 120$~s it is dominated by sickled cells. Excessive distance-transform smoothing suppresses valid markers and leads to under-segmentation, resulting in a rapid increase in counting error, while mild smoothing yields a broad low-error regime. A similar transition is observed for the relative peak-height threshold, with errors remaining low over an intermediate range before increasing sharply as valid peaks are rejected. In contrast, the minimum inter-peak distance exhibits a non-monotonic dependence, reflecting the trade-off between over-segmentation at small values and merged instances at large values. Importantly, within these stable intermediate parameter ranges, the nnU-Net + watershed pipeline consistently achieves substantially lower counting errors than nnU-Net alone for the dominant cell class at each time point, demonstrating robust and improved instance-level counting in dense, overlapping cell populations.

\section*{Reproducible workflow}
All scripts, along with the trained model weights, are publicly available at:\\

\href{https://github.com/nikhil-kadivar/rbc-sickling-dynamics}{\texttt{github.com/nikhil-kadivar/rbc-sickling-dynamics}}.\\

The pipeline requires Python ($\geq$3.9), PyTorch, nnU-Net v2, NumPy, SciPy, scikit-image, OpenCV, and Matplotlib. A minimal \texttt{requirements.txt} and environment setup instructions are provided in the repository to ensure easy installation and reproducibility. A central design objective of this tool is to provide a simple, modular, and fully reproducible workflow that can be executed with minimal setup. 
\begin{verbatim}
# 1) Extract frames 
python extract_frames.py --video /path/to/video.mp4 \
  --all-frames            # OR
  --every-n-frames N      # OR
  --every-sec N

# 2) Run nnU-Net inference to produce PNGs mask files
(0=bg,1=healthy,2=sickle)
python nnunet_infer.py 

# 3) Count, watershed-split, and visualize
python count_and_visualize.py
\end{verbatim}

Each stage of the workflow is fully modular and parameterized, enabling users to adjust the frame sampling rate, output resolution, and processing mode without modifying the underlying code. The pipeline can be executed sequentially across multiple experiments, producing standardized and reproducible sickled ratio estimates alongside visual summaries that provide an effective means for validating the predictions. To maximize computational efficiency, the counting and visualization procedures are parallelized, while nnU-Net inference supports both multi-GPU and multi-CPU configurations. This design ensures that the same codebase can scale seamlessly from a standard laptop to a high-performance computing workstation without additional modification. The resulting outputs include per-frame segmentation overlays, labeled masks, test videos, and tabulated sickling statistics, collectively providing transparent and traceable documentation of the analysis process. Moreover, class definitions and threshold parameters are easily configurable, allowing rapid customization of the workflow for different red blood cell morphologies or related cellular imaging applications. Collectively, these features establish a robust, extensible, and fully reproducible computational framework that converts experimental videos into time-resolved, quantitative maps of sickling dynamics—executable through only a few terminal commands.

\section*{Conclusion}

We developed and validated an automated, modular, and reproducible computational framework to quantify sickling dynamics in densely packed and overlapping RBC suspensions from microfluidic time-lapse videos. The workflow is designed to ensure reproducibility and methodological standardization at every stage—from data preparation to quantitative analysis—by integrating AI-assisted annotation through the Roboflow platform, robust segmentation using a two-dimensional nnU-Net model, and marker-controlled watershed post-processing for accurate instance separation and cell counting in crowded fields of view. Notably, despite being trained on a limited number of annotated frames, the framework achieves high-fidelity segmentation of overlapping cells and exhibits strong agreement with manual quantification, demonstrating robustness to dense cellular arrangements. Experimental validation further confirms that the automatically estimated sickle-cell fractions closely match manual measurements at different levels of hemoglobin modification by the anti-sickling drug osivelotor and reliably capture the temporal progression of sickling. Importantly, the entire pipeline—from frame extraction to quantitative output—can be executed through a minimal three-command interface, enabling straightforward repetition, scalability, and reproducible deployment across computational environments and experimental datasets. By enabling the use of densely packed, overlapping RBC suspensions, this method can  more than double the experimental throughput.

Beyond sickle cell disease, the framework can be readily extended to other biomedical imaging applications involving dynamic changes in object morphology. Its combination of AI-assisted segmentation and quantitative analysis is applicable to studies of RBC deformability in malaria or spherocytosis, as well as investigations of cellular shape evolution, adhesion, or migration in microfluidic and organ-on-chip systems. More broadly, the open-source and modular architecture provides a generalizable foundation for automated, data-driven characterization of cellular biomechanics and therapeutic responses. Looking forward, this framework establishes a foundation for integrating machine learning with experimental biophysics to enable automated, quantitative characterization of cell mechanics. Future extensions may incorporate temporal tracking, three dimensional volumetric analysis, or multimodal imaging data to capture complex morphological alterations under varying biochemical and biomechanical stimuli. By bridging data-driven models with experimental platforms, this approach can accelerate discovery in hematology, mechanobiology, and therapeutic screening, advancing the broader goal of interpretable and reproducible AI in biomedical research.

\section*{Author contributions}
N. K.: Conceptualization, Formal analysis, Methodology, Data curation, Software, Validation, Visualization,  Writing - original draft.\\
G. L.: Conceptualization, Data curation, Formal analysis, Writing - original draft, Visualization, Validation.\\
J. Z.: Performed all experiments, Formal analysis, contributed analytic tools, and wrote the paper.\\
J. M. H.: Provided samples, Formal analysis, and wrote the paper.\\
M. D.: Conceptualization, Project management, Formal analysis, provided experiment-related sources, and wrote the paper.\\
G. K.: Conceptualization, Project management, Formal analysis, Supervision, and wrote the paper.\\
M. X.: Conceptualization, Formal analysis, and wrote the paper.\\

\section*{Conflicts of interest}
All authors declare no conflict of interest.

\section*{Data availability}
All scripts along with trained model are publicly available at
\href{https://github.com/nikhil-kadivar/rbc-sickling-dynamics}{\texttt{github.com/nikhil-kadivar/rbc-sickling-dynamics}}.
\section*{Acknowledgements}

This work was supported by the National Heart, Lung, and Blood Institute of the National Institutes of Health under grant number NIH R01HL154150-06. J.M.H. and M.D. also acknowledge partial support from the NIH under grant number R01HL158102. We thank Dr. John M. Higgins for providing the sickle cell samples. High-performance computing resources were provided by the Center for Computation and Visualization at Brown University. Osivelotor used in this study was provided by Pfizer Inc. through a Pure Compound Grant. Global Blood Therapeutics, the original developer of osivelotor, was wholly acquired by Pfizer Inc. in 2023. We gratefully acknowledge Pfizer Inc. for providing the compound and supporting this research.





\bibliography{sample} 
\bibliographystyle{rsc} 

\end{document}


\section*{Electronic Supplementary Information (ESI)}

\subsection{Effect of watershed post-processing on instance separation and cell counting across suspension densities}

\begin{figure}[!tbp]
    \centering
    \includegraphics[width=0.95\linewidth]{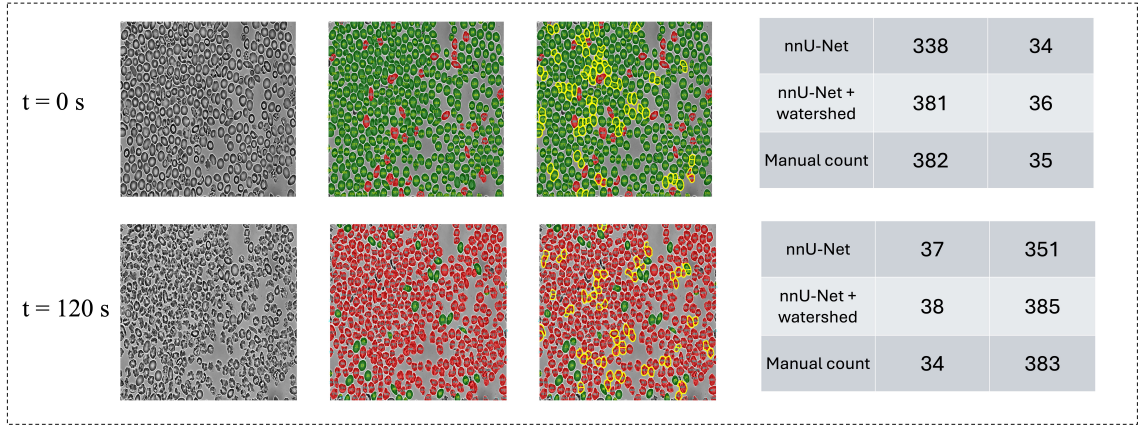}
    \caption{\textbf{Effect of watershed post-processing on instance separation and cell counting across suspension densities.} Representative frames are shown for three cell population densities: Case~1 (\(N = 62\)), Case~2 (\(N = 170\)), and Case~3 (\(N = 417\)), at \(t = 0\)~s and \(t = 120\)~s. Columns display the original image, nnU-Net predictions without watershed post-processing (which tend to merge adjacent or overlapping cells), and nnU-Net predictions with watershed-based instance separation (which splits merged regions into individual cell instances). The accompanying tables report the corresponding healthy and sickled cell counts from each method and from manual counting. In the overlays, green indicates healthy cells, red indicates sickled cells, and yellow highlights regions where overlapping cells are successfully separated by watershed.}
    \label{fig:SI_watershed_importance}
\end{figure}

To evaluate the impact of watershed-based instance separation on quantitative accuracy, we compared results obtained using nnU-Net segmentation with those obtained from nnU-Net followed by watershed post-processing, using manual counting as the reference. As shown in Figure~\ref{fig:SI_watershed_importance}A–C, nnU-Net alone accurately delineates overall cell regions across all suspension densities but increasingly merges adjacent or overlapping RBCs into single connected components as cell density increases from Case~1 (A, N=62) to Case~3 (C, N=417). This merging directly propagates into counting errors, since each merged region is interpreted as a single cell instance, leading to systematic underestimation of both healthy and sickled cell counts. Applying watershed post-processing to the class-specific masks introduces marker-driven boundary propagation that separates merged regions into distinct cell instances, thereby improving agreement with manual counts across all cases. The benefit is modest at low density (Case~1, panel~A) but becomes increasingly pronounced at intermediate and high densities (Case~2, panel~B; Case~3, panel~C), where cell overlap is prevalent. In the highest-density condition (Case~3, panel~C), watershed post-processing substantially reduces discrepancies relative to manual counting at both t = 0~s (predominantly healthy cells) and t = 120~s (predominantly sickled cells). Overall, these results demonstrate that watershed post-processing is a critical step for robust instance-level counting, enabling reliable estimation of sickled-cell fractions under physiologically relevant crowding and overlap conditions.

